\documentclass{article} 
\usepackage{iclr2021_conference,times}

\usepackage{amsmath,amsfonts,bm}

\def\eqref#1{equation~\ref{#1}}

\def\1{\bm{1}}

\DeclareMathAlphabet{\mathsfit}{\encodingdefault}{\sfdefault}{m}{sl}
\SetMathAlphabet{\mathsfit}{bold}{\encodingdefault}{\sfdefault}{bx}{n}

\newcommand{\E}{\mathbb{E}}

\usepackage{hyperref}
\usepackage{url}

\usepackage{graphicx}
\usepackage{subfigure}
\usepackage{wrapfig}

\title{Automatic Goal Generation using Dynamical Distance Learning}

\author{Bharat Prakash, Tinoosh Mohsenin, Tim Oates\\
University of Maryland, Baltimore County\\
Baltimore, MD USA \\
\texttt{\{bhp1,tinoosh,oates\}@umbc.edu} \\
\And
Nicholas Waytowich \\
US Army Research Lab  \\
Aberdeen, MD USA \\
\texttt{\{nicholas.r.waytowich.civ@army.mil} \\
}

\iclrfinalcopy 
\begin{document}

\maketitle

\begin{abstract}
Reinforcement Learning (RL) agents can learn to solve complex sequential decision making tasks by interacting with the environment. However, sample efficiency remains a major challenge. In the field of multi-goal RL, where agents are required to reach multiple goals to solve complex tasks, improving sample efficiency can be especially challenging. On the other hand, humans or other biological agents learn such tasks in a much more strategic way, following a curriculum where tasks are sampled with increasing difficulty level in order to make gradual and efficient learning progress. In this work, we propose a method for automatic goal generation using a dynamical distance function (DDF) in a self-supervised fashion. DDF is a function which predicts the dynamical distance between any two states within a markov decision process (MDP). With this, we generate a curriculum of goals at the appropriate difficulty level to facilitate efficient learning throughout the training process. We evaluate this approach on several goal-conditioned robotic manipulation and navigation tasks, and show improvements in sample efficiency over a baseline method which only uses random goal sampling.

\end{abstract}

\section{Introduction}

Reinforcement Learning agents can learn to solve complex sequential decision making tasks (i.e. MDPs) by interacting with the environment, collecting experience, and learning from that experience. \citep{sutton1998introduction, mnih2015human}. Typically, we need to define a reward function, state and action spaces and the actual learning algorithm. Given these ingredients, the agent can learn to solve MDPs autonomously through trial and error using reinforcement learning (RL). However, a longstanding problem with RL is the relatively poor sample efficiency during learning which can lead to generalization and scalability issues when trying to solve complex tasks. This becomes even more problematic when trying to solve multi-goal tasks with sparse rewards, which are often the type of tasks found in most robotic applications such as manipulation and navigation. 
Humans on the other hand have the ability to learn from very few interactions and also do so with minimal supervision \citep{karni1998acquisition}. There are many reasons for this, but for this paper, we focus specifically on the ability to create and learn from a curriculum. 

Most humans who are trying to learn a new task usually follow a  strategic approach. For example, we may start with easy versions of the task and gradually move to harder ones as we gain more expertise. This concept, known as curriculum learning, has been studied and observed in both humans and animals alike, and is utilized heavily in the fields of psychology and education \citep{krueger2009flexible}. 
This concept has been applied in machine learning as well where we develop a `curriculum' for the agents where the tasks are presented with increasing difficulty levels \citep{bengio2009curriculum}.

\begin{figure*}
  \centering{\includegraphics[width=0.75\textwidth]{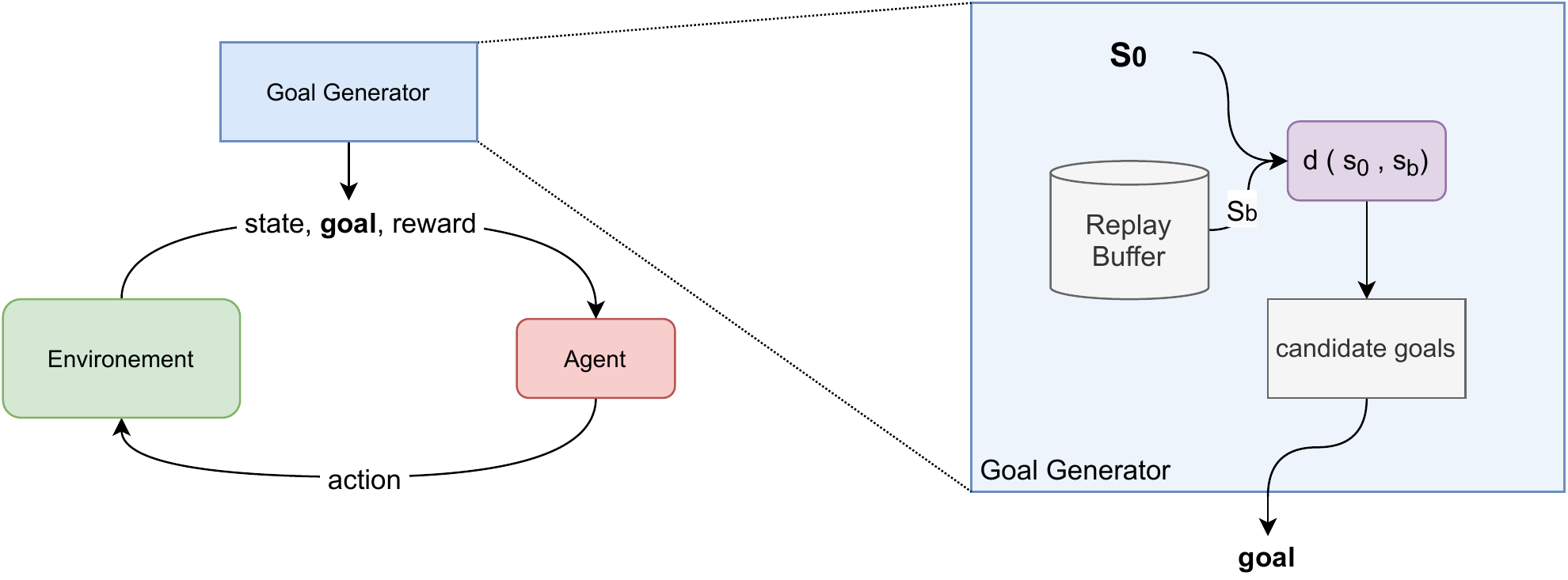}}
  \vspace*{-0.3cm}
  \caption{At every episode, goals are sampled using the goal generator. The distance function $d$ accepts the start state $s_0$ and a random batch of states from the replay buffer and sorts it according to the predicted distance. Candidate goals are then determined based on the required difficulty level(dynamical distance from $s_0$). A goal is sampled from these candidate states.}
    \label{fig:arch}
    \vspace*{-0.3cm}
\end{figure*}

In this work, we focus on automatic curriculum generation to solve goal-conditioned reinforcement learning tasks with sparse rewards. In standard RL, goals (or experience) are sampled randomly without consideration of the relative difficulty of complexity of the goal. Under sparse reward conditions, i.e. only receiving a single reward on task completion, it can be very difficult to learn a policy in a sample efficient way. Curriculum learning, however, can be used in these scenarios to generate goals in a more strategic way depending on the agent's expertise. Generating an appropriate curriculum, however, can be very challenging and labor intensive.  To help address these issues, We propose a technique to develop this curriculum using dynamical distance functions, see Fig. \ref{fig:arch}. Dynamical Distance functions can be learnt in a self supervised way which can approximate distances between 2 states. For evaluation we use 3 goal conditioned robotic manipulations tasks and one Ant-navigation task and show improvements over random goal sampling.

\section{Related Work}

There has been numerous works in developing efficient ways to learn policies with sparse reward functions or even without any reward functions or supervision. \citet{pathak2017curiositydriven} propose a an intrinsic reward called curiosity and \citet{eysenbach2018diversity} show how to learn skills by maximizing an information theoretic objective using a maximum entropy policy. Another interesting approach is latent disagreement as an intrinsic reward by \citet{pathak2019selfsupervised}.Compared to hand engineering a reward function, sparse rewards are easy to design where the agent receives a positive reward only when the goal is reached. Hindsight Experience Replay or HER \citep{andrychowicz2017hindsight} is a method to learn efficiently from sparse rewards and can be combined with an arbitrary off-policy RL algorithm. It has shown to improve sample efficiency and we use this as our backbone algorithm and baseline.

Our work is also inspired from recent work by \citet{hartikainen2019dynamical} where they propose Dynamical Dynamical Distance Learning (DDL). This approach aims to automatically learn dynamical distances, a measure of the expected number of time steps to reach a given goal state from any other state. \citet{tian2020model} also use dynamical distance functions along with model based planning with image observations.

Curriculum Learning has been used in supervised setting for quite some time. \citet{bengio2009curriculum} show that it can help with both speed of convergence as well as accuracy of the final solution. In the reinforcement learning domain \citet{matiisen2019teacher} propose a Teacher-Student framework for automatic curriculum learning where the teacher helps the student to learn a complex task by proposing appropriate sub-tasks. Our method is similar to work by \citep{zhang2020automatic} and \citep{florensa2018automatic}. They both have a separate module which generates goals at appropriate level of difficulty for the agent and this becomes the curriculum.

\section{Preliminaries}

\subsection{Multi-goal Reinforcement Learning}

In standard RL, We have a state space $S$, action space $A$,  and at each time step $t$, the agent takes an action $a_t$ according to some policy $\pi(a_t | s_t)$ that maps from the
current state $s_t$ to a probability distribution over actions. The agent enters a new state $s_{t+1}$ according to a transition function and then receives a reward according to the reward function $r = r(s_t, a_t, s_{s+1})$. The agent is trained to find a policy $\pi$ that  maximises the expected discounted return defined as $$ \E_{s, a, r} [ \sum_{t=0}^{T-1}  \gamma^{t} r_t ] $$. In multi-goal RL, we also have a goal space $G$ as we want to learn policies that can achieve multiple goals. The objective can be changed to include the goal, $ \E_{s, a, r, g} [ \sum_{t=0}^{T-1}]$, here $g$ is sampled from goal space $G$. A simple reward function can be defined as follows $ r(s_t, a, g) = [d(s_t, g) < \epsilon]$ where the agent gets a reward of $0$ when the distance $d$ between the current state and the goal is less than $ \epsilon $, and $\textendash1$ otherwise. The distance $d$ can be Euclidean distance. Multi-goal RL can be formulated as an extension to standard RL where the state space can be $S \times G $ and policy $\pi(a_t | s_t, g)$ that maps state and goal to action.

\begin{figure*}[t]
  \centering

 \subfigure[Fetch Reach]{\includegraphics[scale=0.14]{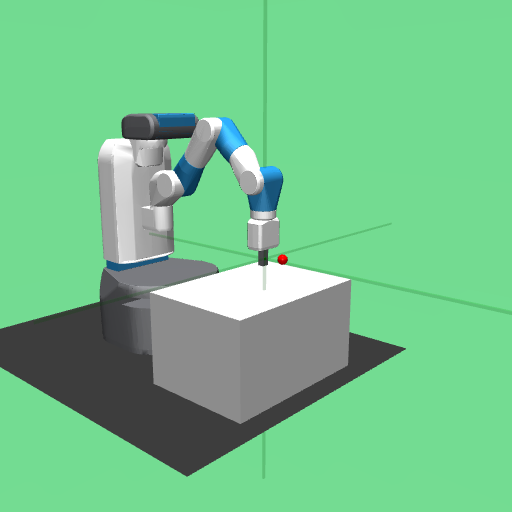}}\quad
  \subfigure[Fetch Push]{\includegraphics[scale=0.14]{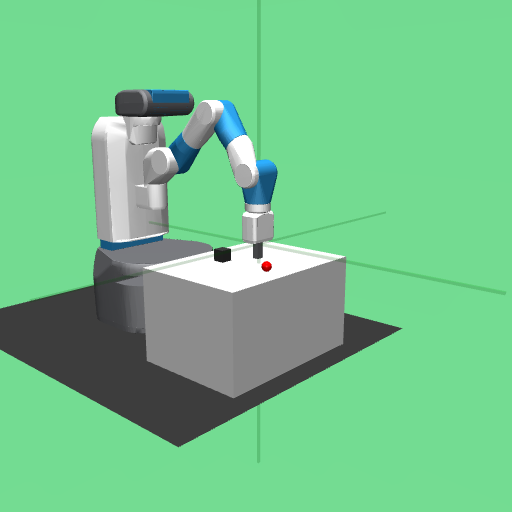}}\quad
  \subfigure[Fetch Pick]{\includegraphics[scale=0.14]{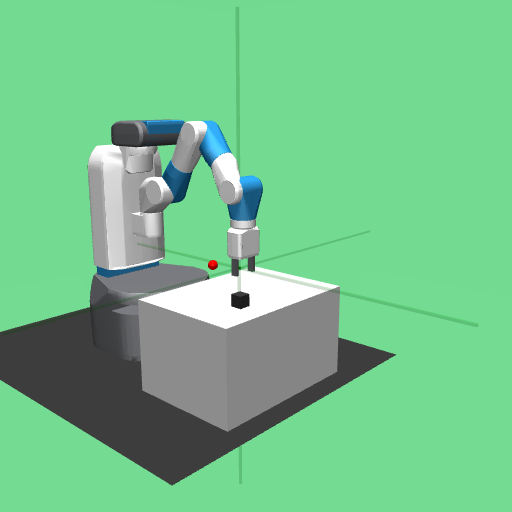}}\quad
  \subfigure[Ant Nav]{\includegraphics[scale=0.14]{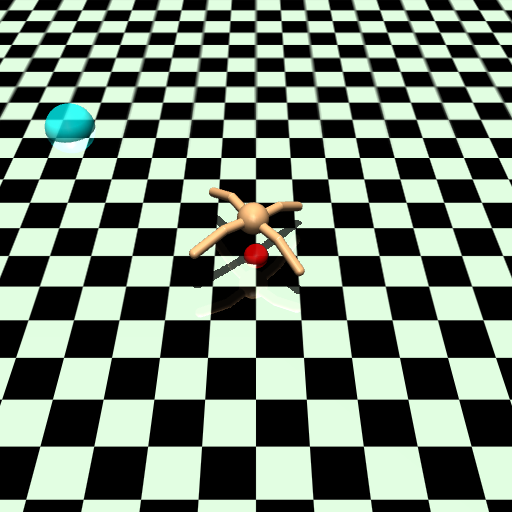}}\quad
  \vspace*{-0.3cm}
 \caption{The 3 goal conditioned robotic manipulations tasks and one Ant-navigation task}
 \vspace*{-0.3cm}
    \label{fig:rewards}
\end{figure*}

\subsection{Dynamical Distance Functions}

Recently, \citep{hartikainen2019dynamical} presented a method called Dynamical Distance Learning (DDL) which learns the expected number of time steps to reach a given goal state from any other state, called dynamical distances. 
The dynamical distance associated with a policy $\pi$, which we write as $d^\pi (s_i, s_j )$, is defined as the
expected number of time steps it took for $\pi$ to reach a state $s_j$ from a state  $s_i$, given that the two were visited in the same episode. Mathematically it can be defined as 
$ d^{\pi}\left(\mathbf{s}, \mathbf{s}^{\prime}\right) = \mathbb{E}_{\tau \sim \pi \mid \mathbf{s}_{i}=\mathbf{s}, \mathbf{s}_{j} = \mathbf{s}^{\prime}, j \geq i}\left[\sum_{t=i}^{j-1} \gamma^{t-i} c\left(\mathbf{s}_{t}, \mathbf{s}_{t+1}\right)\right] $
Here $\tau$ is a trajectory where it passes through $s$ first and then $s^{\prime}$ and $c$ is the cost of moving from $s_i$ to $s_{i+1}$. Both $c$ and $\gamma$ can be set to 1 when there is no supervision and it reduces to just the number of steps needed to reach $s^{\prime}$ from $s$. Training the distance function becomes a regression problem; given two states, it outputs the $distance$ between those states. In order to make the problem simpler and improve the accuracy we convert this into a classification problem. We do this by discretizing the range of possible distances into bins and predict the bin index.


  

 \section{Methods}

In this section we describe the details about our method for automatic goal generation using dynamical distance functions which we use to develop a self-guided curriculum for our learning agent.

Consider a goal space $G$ from which a goal is sampled in multi-goal RL as described in the previous section. Usually, a goal is sampled uniformly at random from this goal space $G$. This means that some of the goals might be very hard to achieve early on in policy development and thus the agent will learn very little from these tasks. Thus, in the early stages of learning a policy, sampling a hard goal is not desirable. When the agent has learnt a decent policy in the later stages, most goals sampled uniformly would be very easy to achieve. As shown in Fig.\ref{fig:arch} we propose an approach where we use a goal generator to sample goals with gradually increasing difficulty levels.

\subsection{Goal Generator}

The goal generator uses a replay buffer from the off-policy RL method and the dynamical distance function (DDF) to produce a set of goal candidates. The DDF is trained at fixed intervals iteratively along with the policy. At the start of every episode, the goal generator receives the initial state $s_0$. It also gets a random batch of states from the replay buffer. We then calculate the distances from the initial state $s_0$ to the sampled batch. As described in the previous section, the distance prediction is treated as a classification problem where distances are arranged in discrete bins. We sample goal states from bins corresponding to furthest distances. These turn out to be just difficult but achievable goals at every stage. 

In the early stages of training, the agent does not have a good policy and explores sub-optimally. The goals sampled by the goal generator using the method described earlier will give the agent easy goals to achieve. As the agent improves, it is able to explore more efficiently and reach further states more often. This means that the replay buffer has better trajectories and DDF is updated along with it. As we continue to do this, we end up sampling goals that are just difficult but achievable at every stage. More details about how we train the distance function is in the appendix.

\begin{figure*}
  \centering{\includegraphics[width=1\textwidth]{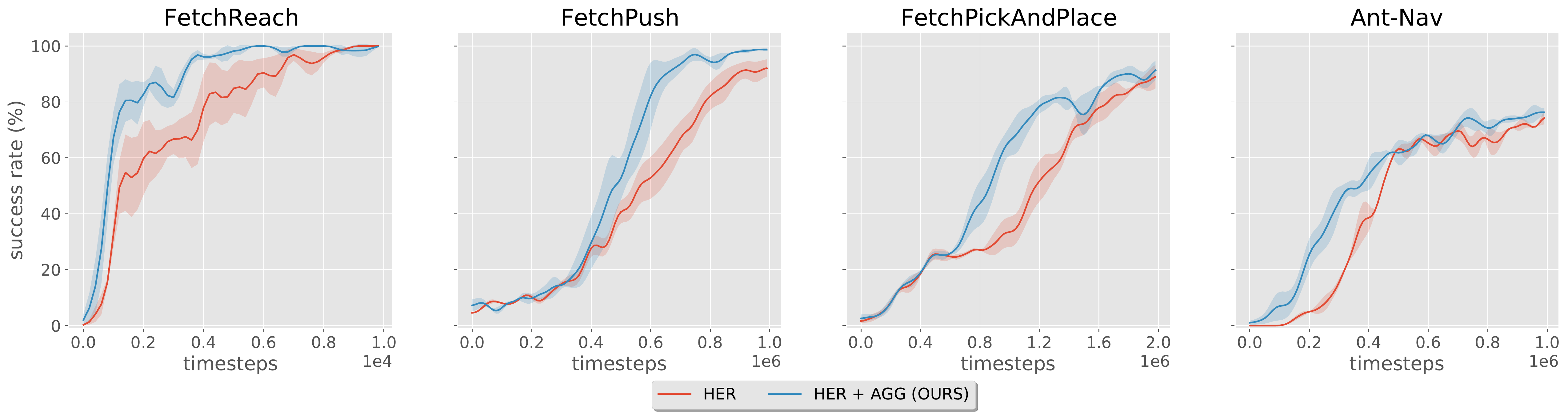}}
  \vspace*{-0.5cm}
  \caption{Learning curves for our experiments. The y-axis shows the success rates achieved by the policy. All plots are averaged over 4 random seeds and the shaded
areas represent standard deviation.}
\vspace*{-0.5cm}
    \label{fig:results}

\end{figure*}

\section{Experiments and Results}

In this section, we describe the experimental setup, training details, and discuss how our method improves performance by creating a curriculum. We test our approach using three different goal-conditioned robotic manipulations tasks and one navigation task all based on the mujoco simulator \citep{todorov2012mujoco} and OpenAI Gym \citep{brockman2016openai}.

\begin{wrapfigure}{R}{0.31\textwidth}
\centering
\includegraphics[width=0.3\textwidth]{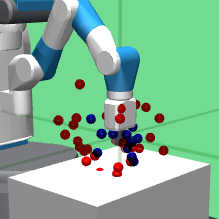}
\vspace*{-0.3cm}
\caption{Visualizing Sampled goals}
\vspace*{-0.3cm}
 \label{fig:curr}
\end{wrapfigure}

The goal generator module is separate from the RL algorithm training. This means that the goal generator can be used along with any off-policy RL method with a replay buffer and a goal-conditioned task. In our case we use hindsight experience replay (HER) with DDPG \citep{lillicrap2015continuous} as the off-policy RL algorithm.
We iteratively train the dynamical distance function in a self-supervised fashion simultaneously along with the policy. The dataset used to train the distance function is the same replay buffer that the agent learns from. More details are in the appendix.

\textbf{Improved sample efficiency}
The results are shown in Figure \ref{fig:rewards}. We can see that our method performs better in all of four environments tested. The final performance of both the baseline and our method do reach the same proficiency for most of the tasks, however,  our method is generally able to converge much faster compared to the baseline indicating improved sample efficiency.

\textbf{Does the DDF help to generate curriculum?}
In order to assess the quality of the curriculum generated, we visualized the goals sampled using the goal generator for the FetchReach task. We sampled goals from the same initial state at different stages of the training process. In Figure \ref{fig:curr} we show the goals sampled in the early stages of the learning in blue. As can be seen, the goals at this stage are much closer to the initial arm position and are easy goals. The goals sampled during the later stages of the training are shown in red. Here we can see that these goals are much more challenging and they are presented only during the later stages when the agent has a decent policy. This shows that the goal generator samples goals with gradually increasing difficulty level.

\section{Conclusion}
In this work we present a method for automatic goal generation which creates a self-guided curriculum for our learning agent using DDF. We test this on several robotic manipulation and navigation tasks and show improvements in sample efficiency compared to an HER baseline. Our method does not require any supervision to create the curriculum and can potentially be extended to more complex sparse reward tasks with image based observations and real world robots.

\bibliography{rl}
\bibliographystyle{iclr2021_conference}

\raggedbottom

\pagebreak
\appendix
\section{Appendix}

\subsection{Training the Dynamical Distance Function}

In order to train the DDF, we use the replay buffer using in the off-policy RL alogrothm. The replay buffer stores trajectories of state, action, next state tuples. We use this to create a dataset of pairs of states as input and distance as the label. Distance this this case is just the number of time steps between states. 

In order to make the problem simpler and improve the accuracy we wanted to convert this into a classification problem. We do this by discretizing the distances into bins and predict the bin index. Eg. distances 1-10 belong to bin 1, 11-20 belong to bin 2 and so on. We then can train the network using a cross entropy loss as we usually do in classification problems.

Instead of a linear discretization, we decided to create bins by placing distances on the log scale (geometric progression). For instance, lets say the maximum length of the trajectory is 50 and we want 5 bins. Distance of 1 belongs to bin 1, 2-5 belongs to bin 2, 6-11 belong to bin 3, 11-21 belong to bin 4, 21-50 belongs to bin 5 and so on. This helps balance the dataset and we found that it gave us better results. In our goal generator we, given 2 states, we predict these bins. Now we can query states corresponding to bin 5 and they would be the furthest states.

The DDF is re-trained at fixed intervals after every 50000 to 100000 steps depending on the environment. The replay buffer size for the HER algorithm is set to bigger number. We use $1e6$ in our experiments. But for the training the DDF we use a subset of the replay buffer with more recent trajectories. We choose the recent $1e5$ steps to train the DDF.

\subsection{Hindsight Experience Replay}
 Hindsight Experience Replay or HER \citep{andrychowicz2017hindsight} is a method to learn efficiently from sparse rewards and can be combined with an arbitrary off-policy RL algorithm. Episodes are stored in a replay buffer along with the achieved and desired goals at each transition. HER works by adding additional transitions with different goals. Usually this goal is a state which was reached later in the same episode. Trajectories can be replayed with arbitrary goals multiple times and HER works with any off-policy optimization.

\subsection{Visualizing Goal generation for Fetch Reach}
We visualized the goals sampled using the goal generator for the FetchReach task. We sampled goals from the same initial state at different stages of the training process.
In the early stage, the goals are much closer to the initial arm position. The goal generator gradually samples more difficult goals as training progresses.  

\begin{figure}[h]
  \centering

 \subfigure[Easy Goals]{\includegraphics[scale=0.24]{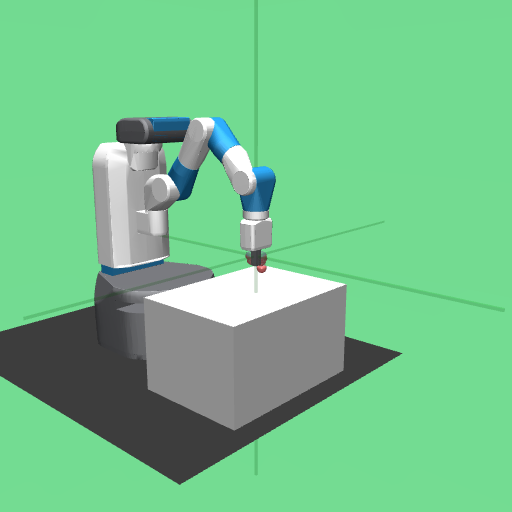}}\quad
  \subfigure[Medium Goals]{\includegraphics[scale=0.24]{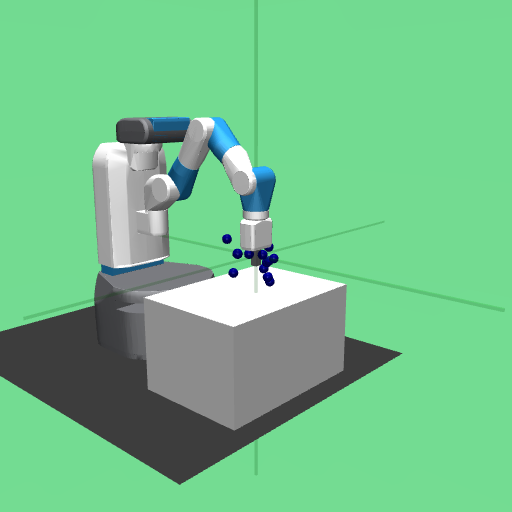}}\quad
  \subfigure[Hard Goals]{\includegraphics[scale=0.24]{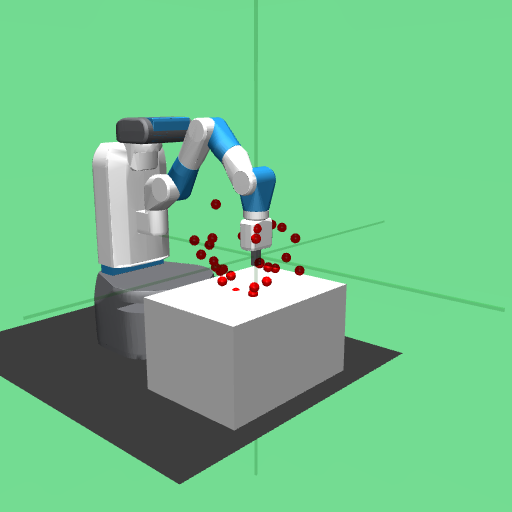}}\quad
  \vspace*{-0.3cm}
 \caption{Goal generator creates a curriculum}
 \vspace*{-0.3cm}
    \label{fig:goals}
\end{figure}

\end{document}